%% file: main.tex
\documentclass[sigconf]{acmart}

\usepackage{latexsym}
\usepackage{tikz-cd} 
\usepackage{amsthm}
\usepackage{pgfplots}
\usetikzlibrary{patterns}
\usetikzlibrary{plotmarks}
\pgfplotsset{compat=1.11,
    /pgfplots/ybar legend/.style={
        /pgfplots/legend image code/.code={%
            \draw[##1,/tikz/.cd,bar width=3pt,yshift=-0.2em,bar shift=0pt]
            plot coordinates {(0cm,0.8em)};
        },
    },
}

\usepackage{microtype}
\usepackage{caption}
\usepackage{balance}
\usepackage{graphicx}
\usepackage{subcaption}
\usepackage[inline]{enumitem}
\usepackage{booktabs}
\usepackage{arydshln}
\usepackage{multirow}
\usepackage{comment}

\newcommand{\term}[1]{{\small\tt #1}}

\newcommand{\squishlist}{
 \begin{list}{$\bullet$}
  { \setlength{\itemsep}{0pt}
     \setlength{\parsep}{1pt}
     \setlength{\topsep}{1pt}
     \setlength{\partopsep}{0pt}
     \setlength{\leftmargin}{1.5em}
     \setlength{\labelwidth}{1em}
     \setlength{\labelsep}{0.5em} } }
 \newcommand{\squishend}{\end{list}}

\newcommand{\triple}[1]{\emph{$\langle$#1$\rangle$}}

\newcommand{\code}[1]{\texttt{\textbf{{\small #1}}}}

\newcommand{\sr}[1]{{\textcolor{violet}{SR: #1}}}

\newcommand\ourkb{\textsc{Ascent}}

\renewcommand{\paragraph}[1]{\smallskip\noindent\textbf{#1.\mbox{\ \ }}}

\title{Advanced Semantics for Commonsense Knowledge Extraction}

\author{Tuan-Phong Nguyen}
 \affiliation{
     \institution{Max Planck Institute for Informatics}
 }
 \email{tuanphong@mpi-inf.mpg.de}

 \author{Simon Razniewski}
 \affiliation{
     \institution{Max Planck Institute for Informatics}
 }
 \email{srazniew@mpi-inf.mpg.de}

 \author{Gerhard Weikum}
 \affiliation{
     \institution{Max Planck Institute for Informatics}
 }
 \email{weikum@mpi-inf.mpg.de}

\copyrightyear{2021}
\acmYear{2021}
\setcopyright{iw3c2w3}
\acmConference[WWW '21]{Proceedings of the Web Conference 2021}{April 19--23, 2021}{Ljubljana, Slovenia}
\acmBooktitle{Proceedings of the Web Conference 2021 (WWW '21), April 19--23, 2021, Ljubljana, Slovenia}
\acmPrice{}
\acmDOI{10.1145/3442381.3449827}
\acmISBN{978-1-4503-8312-7/21/04}

\setcopyright{iw3c2w3}

\begin{document}
\begin{abstract}
Commonsense knowledge (CSK) about concepts and their properties is 
useful
for AI applications such as robust chatbots. 
Prior works like ConceptNet, TupleKB and others 
compiled large CSK collections,
but 
are restricted in their 
expressiveness to
subject-predicate-object (SPO) triples with simple
concepts for S
and
monolithic strings for P and O.
Also, these projects have either prioritized
precision or recall, but hardly reconcile these complementary goals.
This paper presents a methodology, called \ourkb{}, to automatically build a large-scale 
knowledge base (KB) of CSK assertions, with advanced expressiveness
and both better precision and recall than
prior works. 
\ourkb{} goes beyond triples by capturing
composite concepts with subgroups and aspects, and by
refining assertions with semantic facets.
The latter are important to 
express 
temporal and spatial validity of assertions
and 
further qualifiers.
%
\ourkb{} combines open information extraction with judicious cleaning 
using
language models.
Intrinsic evaluation
shows the superior size and quality of the
\ourkb{} KB, and an extrinsic evaluation for
QA-support tasks 
underlines the benefits of \ourkb{}.
A web interface, data and code can be found at \url{https://ascent.mpi-inf.mpg.de/}.
%
\end{abstract}

\maketitle


\input{sections/1-Introduction}
\input{sections/2-Related-Work}

\input{sections/3-Model-and-Architecture}

\input{sections/4-Methodology}

\input{sections/5-Experiments}

\input{sections/6-Conclusion}

\medskip

\paragraph{Acknowledgment}
We are thankful to Kyle Richardson for suggestions on extrinsic comparisons of CSKBs.



\balance

\bibliographystyle{acl_natbib}
\bibliography{refs}


\end{document}

%% file: sections/1-Introduction.tex
\section{Introduction}
\paragraph{Motivation} 
Commonsense knowledge (CSK) is a long-standing 
goal
of AI~\cite{mccarthy1960programs,feigenbaum1984knowledge,lenat1995cyc}:
equip machines with structured knowledge
about everyday concepts and their properties
(e.g., elephants are big and eat plants, 
buses carry passengers and drive on roads)
and about
typical human behavior
and emotions 
(e.g., children love visiting zoos, 
children enter buses to go to school).
%
In recent years, research on automatic acquisition of CSK assertions has been
greatly advanced and several commonsense knowledge bases (CSKBs) of considerable size have been
constructed (see, e.g., \cite{conceptnet,webchild,mishra2017domain,quasimodo}). 
Use cases for CSK 
include particularly 
language-centric tasks such as question answering and conversational systems 
(see, e.g., \cite{lin2017reasoning,lin2019kagnet,xia2019incorporating}).

%
\textit{Examples}: 
Question-answering systems often need CSK as background knowledge for robust answers.
For example, when a child asks ``Which zoos have habitats for T-Rex dinosaurs?'', 
the system should point out that 
i) dinosaurs are extinct, and ii) can be seen
in museums, not in zoos.
Dialogue systems should not just generate plausible utterances from a language model, but 
should be situative, understand metaphors and implicit contexts and avoid blunders. For example, when a user says
``tigers will soon join the dinosaurs'', the
machine should understand that this refers to
an endangered species rather than alive tigers 
invading
museums.
%

The goal of this paper is to advance 
the automatic acquisition of CSK assertions from online contents
better expressiveness, higher precision
and wider coverage. 



\vspace{-6pt}

\paragraph{State of the Art and its Limitations} 
Large KBs like DBpedia, Wikidata or Yago largely focus on encyclopedic knowledge on individual entities
like people, places etc., and 
and are very sparse on general concepts~\cite{ilievski2020commonsense}. 
Notable projects that focus on CSK include ConceptNet~\cite{conceptnet}, WebChild~\cite{webchild}, Aristo TupleKB~\cite{mishra2017domain} and Quasimodo~\cite{quasimodo}. 
They are all based on SPO triples as knowledge representation and have major shortcomings:
\squishlist
    \item \emph{Expressiveness for S:} 
    As subjects, prior CSKBs 
    strongly focus on simple concepts expressed
    by single nouns (e.g., elephant, car, trunk).
    This misses semantic refinements (e.g., diesel car vs. electric car) that lead to different properties (e.g., polluting vs. green), and is
    also prone to word-sense disambiguation problems (e.g., elephant trunk vs. car trunk).
    Even when CSK acquisition considers multi-word phrases, it still lacks the awareness of
    semantic relations among concepts.
    Hypernymy lexicons like WordNet or Wiktionary
    are also very sparse on multi-word concepts.
    With these limitations, word-sense disambiguation does not work robustly; prior attempts showed mixed results at best 
    (e.g., \cite{webchild,mishra2017domain}).
    \item \emph{Expressiveness for P and O:} 
Predicates and objects are treated as monolithic strings, such as
\squishlist
\item[o] A1: buses, [used for], [transporting people];
\item[o] A2: buses, [used for], [bringing children to school];
\item[o] A3: buses, [carry], [passengers];
\item[o] A4: buses, [drop], [visitors at the zoo on the weekend].
\squishend
This misses the 
equivalence of
assertions A1 and A3, and is unable to capture
the semantic relation between A1 and A2, namely,
A2 refining A1. Finally, the spatial facets of
A2 and A4 are cluttered into unrelated strings,
and the temporal facet in A4 is not explicit either. 
The alternative of restricting P to a small number
of pre-specified predicates (e.g., \cite{conceptnet,webchild})
and O to very short phrases comes at the cost of much lower coverage. 
    \item \emph{Quality of CSK assertions:} 
    Some of the major CSKBs have prioritized precision (i.e., the validity of the assertions) but have fairly limited coverage
    (e.g., \cite{conceptnet,mishra2017domain}.
    Others have wider coverage but include 
    many noisy if not implausible assertions
    (e.g., \cite{webchild,quasimodo}).
    Very few have paid attention to the saliency of assertions, i.e., the degree to which statements are common knowledge, as opposed
    to merely capturing many assertions. Projects along these lines (e.g., \cite{conceptnet,atomic}) 
    fall short in coverage, though. 
\squishend
\ourkb{} aims to overcome these limitations of prior works, while retaining their positive characteristics. In particular, we aim to 
reconcile 
high precision 
with
wide coverage
and saliency.
Like \cite{mishra2017domain,quasimodo}, we aim to acquire open assertions (as opposed to pre-specified predicates only), but 
strive for more expressive representations
by 
refining subjects and capturing semantic facets of assertions.
\paragraph{Approach}
We present the \ourkb{} method for
acquiring CSK assertions with
advanced semantics, from
web contents.
\ourkb{} operates in three phases: 
(i) 
source discovery, 
(ii) open information extraction (OIE), 
(iii) 
automatic 
consolidation.
In the first phase, \ourkb{} generates
search queries for a given target concept such as ``star'' to retrieve relevant 
pages. The queries include 
hypernyms
from lexicons such as WordNet,
this  way
covers
different meanings of ``star''
while
distinguishing results for ``star (celebrity)''
(with hypernym ``human'') vs. ``star (celestial body)'' (with hypernym ``natural object''). 
Results are further scrutinized by
comparing, via embedding similarity,
against the respective Wikipedia articles.
In the second phase, \ourkb{} collects OIE-style tuples by carefully designed dependency-parse-based rules, taking into account assertions for subgroups and aspects of target subjects, and increasing recall by co-reference resolution. 
The extractors use cues from prepositional phrases to detect semantic facets, and use supervised classification for eight facet types. 
Finally, in the 
consolidation phase,
assertions are iteratively grouped and semantically organized by an efficient combination of filtering based on fast word2vec similarity, and classification based on a fine-tuned RoBERTa language model.

We ran \ourkb{} for 10,000 
frequently used
concepts as target subjects.
The resulting CSKB significantly 
outperforms automatically built state-of-the-art CSK 
collections in salience and recall.
In addition, we performed
an extrinsic evaluation in which commonsense knowledge was used to support language models in question answering. 
 \ourkb{} significantly outperformed
 language models without context, and was consistently among the top-scoring KBs in this evaluation.

\paragraph{Contributions}
Salient contributions of this work are:
\squishlist
    \item introducing an expressive model for commonsense knowledge with advanced semantics, with subgroups of subjects and faceted assertions as first-class citizens;
    \item developing a fully automated methodology for populating the model with high-quality CSK assertions by extraction from web contents;
    \item constructing a large CSKB for 10,000 important concepts.
\squishend
%
A web interface to the \ourkb{} KB, along with downloadable data and code is available at \url{https://ascent.mpi-inf.mpg.de/}.

%% file: sections/2-Related-Work.tex
\section{Related Work}

\paragraph{Commonsense knowledge bases (CSKBs)}
CSK acquistion has a long tradition in AI
(e.g., \cite{lenat1995cyc,singh2002open,liu2004conceptnet,DBLP:conf/aaai/GordonDS10}). 
A few projects have constructed large-scale collections that are
publicly available.
ConceptNet~\cite{conceptnet} is the most prominent project on CSK acquisition. Relying 
mostly
on human crowdsourcing, it contains highly salient information 
for a small number of pre-specified predicates
(isa/type, part-whole, used for, capable of, location of, plus lexical relations such as synonymy, etymology, derived terms etc.),
and this CSKB is most widely used. 
However, it has limited coverage on
many concepts, and its ranking of assertions, based on the number of crowdsourcing inputs, is very spares and unable to discriminate salient properties against atypical or exotic ones
(e.g., listing trees, gardens and the bible as locations of snakes, with similar scores).
ConceptNet does not properly disambiguate concepts, leading to incorrect assertion chains like \triple{elephant, hasPart, trunk}; \triple{trunk, locationOf, spare tire}.

WebChild~\cite{webchild}, TupleKB~\cite{mishra2017domain} and Quasimodo~\cite{quasimodo} devised
fully automated methods for CSKB construction. They use judiciously selected text corpora (incl. book n-grams, image tags, QA forums) to extract large amounts of SPO triples. 
WebChild builds on hand-crafted extraction patterns, and TupleKB and Quasimodo rely on open information extraction with subsequent cleaning.
All three are limited to SPO triples.

Recently, TransOMCS \cite{DBLP:conf/ijcai/ZhangKSR20}
has harnessed statistics about preferential attachment to convert
a large linguistic collection of patterns into a CSKB of SPO triples
with a pre-specified set of predicates.
It uses Transformer-based neural learning for
plausibility scoring.

We adopt the idea of using search engines for source discovery and open information extraction (OIE).
Our novelty for source discovery 
lies in 
generating better focused queries and
scrutinizing candidate documents against reference Wikipedia articles. 
For extraction, we extend OIE to
capture expressive facets and also multi-word compounds as subjects. Multi-word compounds enable a higher recall on salient assertions, as well as avoiding common disambiguation errors.

\paragraph{Taxonomy and meronymy induction}
The organization of concepts in terms of subclass and part-whole relationships, termed hypernymy and meronymy,
has
received great attention in NLP
and web mining (e.g.,
\cite{DBLP:conf/www/EtzioniCDKPSSWY04,DBLP:conf/acl/SnowJN06,DBLP:journals/coling/GirjuBM06,DBLP:conf/acl/PantelP06,DBLP:conf/acl/PascaD08,DBLP:journals/ai/PonzettoS11,DBLP:conf/sigmod/WuLWZ12,HertlingPaulheim:ISWC2017}).
The hand-crafted WordNet lexicon \cite{wordnet}
organizes over 100k synonym sets with respect to these relationships, although meronymy is sparsely populated.

Recent methods for large-scale taxonomy induction from web sources include 
WebIsADB \cite{seitner2016large,HertlingPaulheim:ISWC2017} building on Hearst patterns and other techniques,
and the industrial GIANT ontology 
\cite{DBLP:conf/sigmod/LiuGNLWWX20}
based on neural learning from user-action logs and other sources.

Meronymy induction at large scale
has been addressed by \cite{tandon2016commonsense,haspartkb,bhakthavatsalam2020genericskb} 
with pre-specified and automatically learned patterns for refined relations
like physical-part-of, member-of and
substance-of.

Our approach includes relations of both kinds, 
by extracting knowledge about
salient subgroups and aspects 
of subjects. 
In contrast to typical taxonomies and part-whole collections, our subgroups include many multi-word phrases: composite noun phrases 
(e.g., ``circus lion'', ``lion pride'') and
adjectival and verbal phrases
(e.g., ``male lion'', ``roaring lion'').
Aspects cover additional refinements of subjects that do not fall under taxonomy or meronymy
(e.g., ``lion habitat'' or ``lion's prey'').

\paragraph{Expressive knowledge representation and extraction}
Modalities such as \term{always, often, rarely, never}
have a long tradition in AI research
(e.g., \cite{gabbay2003many}), based on various kinds of modal logics or semantic frame representations, and semantic web formalisms can capture context using e.g., RDF* or reification~\cite{hoganetal}.
While such expressive knowledge representations have been around for decades, there has hardly been any work that populated KBs with such refined models,
notable exceptions being the 
Knext project
\cite{schubert2002can} at small scale,
and OntoSenticNet \cite{DBLP:journals/expert/DragoniPC18} with focus on affective valence annotations.

Other projects have pursued different kinds of contextualizations for CSK extraction, notably \cite{zhang2017ordinal}, which scored natural language sentences on an ordinal scale covering the spectrum \emph{very  likely, likely, plausible, technically possible} and \emph{impossible}, \citet{chen-etal-2020-uncertain}
with probabilistic scores, and the Dice project
\cite{chalier2020joint} which ranked assertions along 
the dimensions of plausibility, typicality and saliency. 

Semantic role labelling (SRL) is 
a representation and methodology
where sentences are mapped onto
frames (often for certain types of events)
and respective slots 
(e.g., agent, participant, instrument)
are filled with values extracted from the input text
\cite{DBLP:series/synthesis/2010Palmer,clarke2012nlp,semanticrolelabelling}. 
Recently, this paradigm has been extended towards facet-based open information extraction, where extracted tuples
are qualified with semantic facets like location and mode~\cite{stuffie,graphene}.
\ourkb{} builds on this general approach, but extends it in various ways geared for the case of CSK:
focusing on specifically relevant facets, refining subjects by subgroups and aspects, and aiming to reconcile precision and coverage for concepts as target subjects.



\paragraph{Pre-trained language models}
Recently, there has been great progress on
pre-trained language models (LMs) like BERT and GPT
\cite{devlin2019bert,DBLP:journals/corr/abs-2005-14165}.
%
In \ourkb{} we make use of such language models, utilizing them to cluster semantically similar phrases in order to reduce redundancy and group related assertions. We also use LMs in the extrinsic evaluation for question answering, showing that priming LMs with structured knowledge from CSKBs can greatly improve performance (cf. also \cite{petroni2020context}).

%% file: sections/3-Model-and-Architecture.tex

\section{Model and Architecture}

\subsection{Knowledge Model}

Existing CSKBs typically follow a triple-based data model, where subjects are linked via predicate phrases to 
object words or phrases.
Typical examples, from ConceptNet,
are
\triple{bus, usedFor, travel}
and
\triple{bus, usedFor, not taking the subway}.
Few projects \cite{webchild,mishra2017domain} have attempted to sharpen such assertions by word sense disambiguation (WSD) \cite{DBLP:journals/csur/Navigli09}, distinguishing, for example, buses on the road from computer buses.
Likewise, only few projects
\cite{DBLP:conf/aaaifs/GordonS10,zhang2017ordinal,quasimodo,chalier2020joint}, 
have tried to identify salient assertions against correct ones that are unspecific, atypical or even misleading (e.g., buses used for avoiding the subway or used for enjoying the scenery).
We extend this prevalent paradigm in two major ways.

\paragraph{Expressive subjects} 
CSK acquisition starts by collecting 
assertions for target subjects, which are usually single nouns. 
This has two handicaps: 1) it conflates different meanings for the same word, and 2) it misses out on refinements and variants of word senses.
While word sense disambiguation (WSD) has been tried to overcome the first issue
\cite{webchild,mishra2017domain},
it has been inherently limited because
the underlying word-sense lexicons,
like WordNet and Wiktionary, mostly restrict themselves to single nouns.
For example, phrases like ``city bus'' or ``tourist bus'' are not present at all. 


Our approach to rectify this problem is twofold:
\squishlist
\item First, our source discovery method combines the target subject with an informative hypernym (using WordNet,
applied to single nouns or head words in phrases). For example, instead of searching with the semantically overloaded word ``bus'', we generate queries ``bus public transport'' and
``bus network topology'' to disentangle the different senses.
\item Second, when extracting candidates for assertions from the retrieved web pages, we capture also multi-word phrases as candidates for refined subjects, such as ``school bus'', ``city bus'', ``tourist bus'', ``circus elephant'', ``elephant cow'', ``domesticated elephant'', etc.
This way, we can acquire {\em isa}-like refinements, to create {\em subgroups} of broader subjects, and also other kinds of {\em aspects} that are relevant to the general concept.
An example for the latter would be ``bus driver'' or, for the target subject ``elephant'', phrases such as ``elephant tusk'', ``elephant habitat'' or ``elephant keeper''.
\squishend


Our notion of {\em subgroups}
can be thought of as an inverse {\em isa} relation. It goes beyond traditional taxonomies by better coverage of multi-word composites
(e.g., ``circus elephant'').
This allows us to better represent specialized assertions such as 
\triple{circus elephants, catch, balls}.

Our notion of {\em aspects} includes
part-whole relations (partOf, memberOf, substanceOf) \cite{DBLP:journals/coling/GirjuBM06,shwartz2018olive,tandon2016commonsense,haspartkb}, but also further aspects that do not fall under the
themes of hypernymy or meronymy.
Examples are ``elephant habitat'',
``bus accident'', etc.
Note that, unlike single nouns, such compound phrases are rarely ambiguous, so we have crisp concepts without the need for explicit WSD.

\paragraph{Semantic facets} For CSK, assertion 
validity
depends often on specific temporal and spatial circumstances, e.g., elephants scare away lions only in Africa, or bathe in rivers only during daytime. 
Furthermore, assertions often
become crisper by contextualization in terms of causes/effects and instruments
(e.g., children ride the bus \dots to go to school, circus elephants catch balls \dots with their trunks). 

\begin{figure*}[t]
    \centering
    \includegraphics[width=0.65\textwidth]{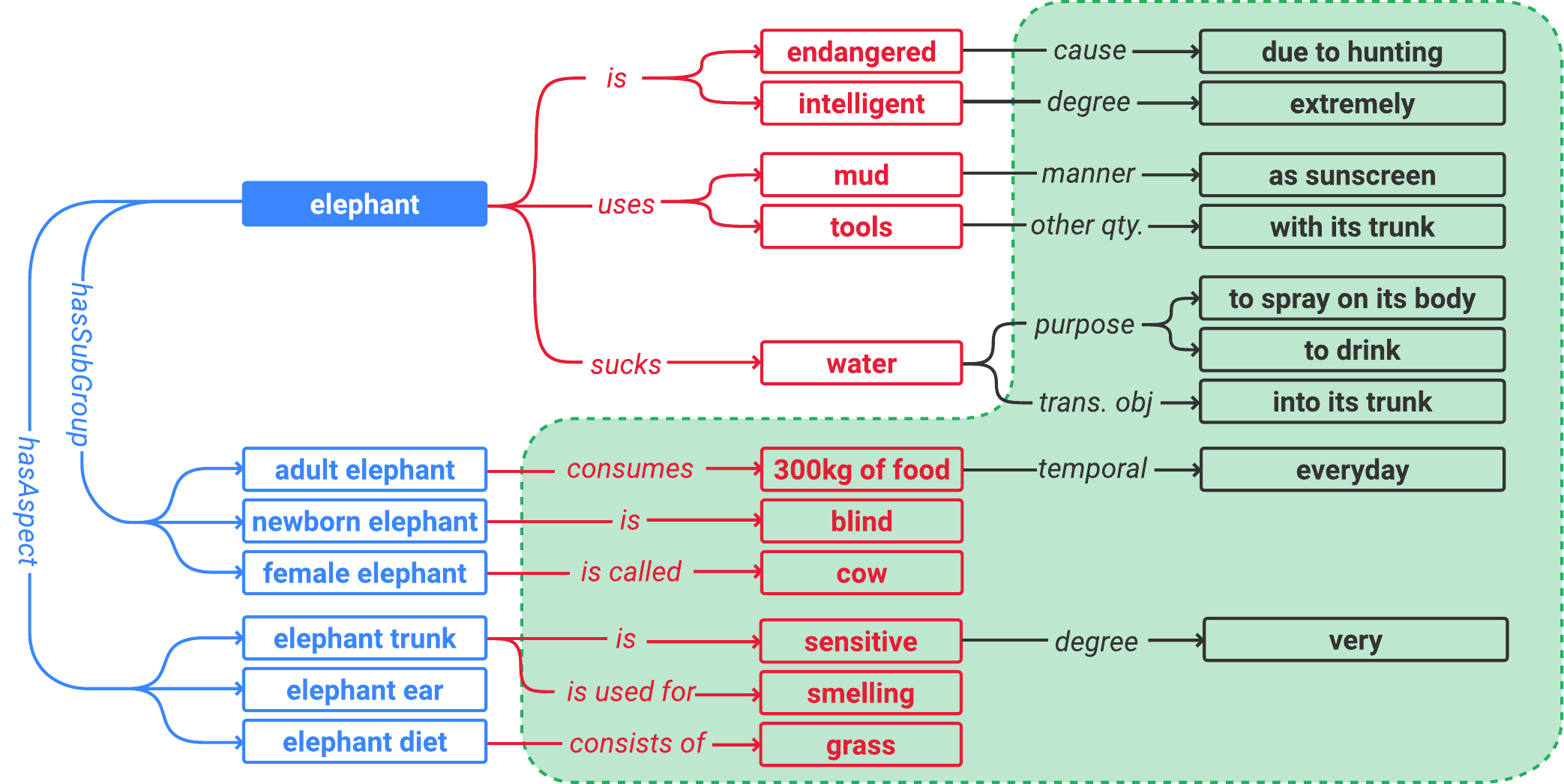}
    \caption{Example of \ourkb's knowledge for the concept \emph{elephant}. The data model of traditional CSKBs like ConceptNet is restricted to assertions outside the green box.}
    \label{fig:examples}
\end{figure*}

To incorporate such information into an expressive model, we choose to contextualize subject-predicate-object triples  with semantic {\em facets}. 
To this end, we build on ideas from research on semantic role labeling
(SRL) \cite{DBLP:series/synthesis/2010Palmer,clarke2012nlp,semanticrolelabelling}.
This line of research has originally been devised to fill  hand-crafted frames (e.g., purchase) with values for frame-specific roles (e.g., buyer, goods, price etc.).
We start with a set of 35 labels proposed in~\cite{stuffie}, a combination of those in the Illinois Curator SRL~\cite{clarke2012nlp} and 22 hand-crafted ones derived from an analysis of semantic roles of prepositions in Wiktionary (\url{https://en.wiktionary.org/}).
As many of these are very special, we condense them into eight widely useful roles that are of relevance for CSK: 4 that qualify the validity of assertions (degree, location, temporal, other-quality), and 4 that capture other dimensions of context (cause, manner, purpose, transitive objects).

These design considerations lead us to the following knowledge model.

\vspace*{0.2cm}
\noindent{\bf Definition [Commonsense Assertion]:}\\
Let $C_0$ be a set of primary concepts of interest, which could be manually defined or taken from a dictionary. \\
%
Subjects for assertions include all $s_0 \in C_0$ as well as judiciously selected
multi-word phrases that contain some $s_0$.\\
Subjects are interrelated by 
{\em subgroup} and {\em aspect}
relations: each $s_0$ can be refined
by a set of subgroup subjects denoted
$sg({s_0})$, and by a set of aspect subjects denoted $asp({s_0})$.
The overall set of subjects is
$C := C_0 \cup sg_{C_0} \cup asp_{C_0}$.\\
A commonsense assertion for $s \in C$ is a quadruple \triple{s, p, o, F}
with single-noun or noun-phrase subject $s$, short phrases for predicate $p$
and object $o$ and a set $F$ of 
semantic facets.
Each facet $(k,v) \in F$
is a key-value pair with one of eight possible keys $k$ and a short phrase as $v$. 
Note that a single assertion can have multiple key-value pairs with the same key (e.g., different spatial phrases).
\hspace*{0pt}\hfill $\qed$
\vspace*{0.1cm}

An example of assertions for $s_0=\text{elephant}$ is shown in Fig.~\ref{fig:examples}.

\subsection{Extraction Architecture}

\paragraph{Design considerations}
CSK collection has three major design points: (i) the choice of sources, (ii) the choice of the extraction techniques, and (iii) the choice of cleaning or consoliding the extracting candidate assertions.

As \textit{sources},
most prior works carefully selected high-quality input sources, including book n-grams \cite{webchild}, concept definitions in encyclopedic sources, and school text corpora about science \cite{clark2018think}.  
These are often a limiting factor
in the KB coverage. Moreover, even
seemingly clean texts like book n-grams come with a surprisingly high level of noise and bias (cf. \cite{DBLP:conf/cikm/GordonD13}).
Focused queries for retrieving suitable web pages were used by \cite{mishra2017domain}, but the 
query formulations required non-negligible effort.
Query auto-completion and question-answering forums were tapped by Quasimodo~\cite{quasimodo}.
While this gave access to highly salient assertions, it was, at the same time,
adversely affected by heavily biased and
sensational contents
(e.g., search-engine auto-completion for 
``snakes eat'' suggesting ``\dots themselves'' and ``\dots children'').
In \ourkb{} we opt for using search engines for wide coverage, and devise
techniques for quality assurance.

For the \textit{extraction techniques}, choices range from co-occurrence- and pattern-based methods (e.g.,~\cite{elazar2019large})
and open information extraction (OIE)
(e.g.,~\cite{mishra2017domain,quasimodo})
to supervised learning for classification and sequence tagging.  
Co-occurrence works well for a few pre-specified, clearly distinguished predicates, using distant seeds. Supervised extractors require training data for each predicate, and thus have the same limitation. 
Recent approaches, therefore, prefer
OIE techniques, and the \ourkb{}
extractors follow this trend, too.

For \textit{knowledge consolidation}, early approaches simply kept all assertions from the ingest process (e.g., crowdsourcing \cite{conceptnet}), whereas recent projects employed supervised classifiers or rankers for cleaning \cite{mishra2017domain,quasimodo,chalier2020joint}, and 
also limited forms of clustering~\cite{mishra2017domain,quasimodo} for canonicalization (taming semantic redundancy). 
In \ourkb{}, the careful source selection already eliminates certain kinds of noise, rendering extraction frequency statistics a much better signal than in earlier works.
Therefore, we focus on reinforcing these signals for consolidation,
based on clustering with contextual language models for informative similarity measures.

\begin{figure*}[t]
    \centering
    \includegraphics[width=0.65\textwidth]{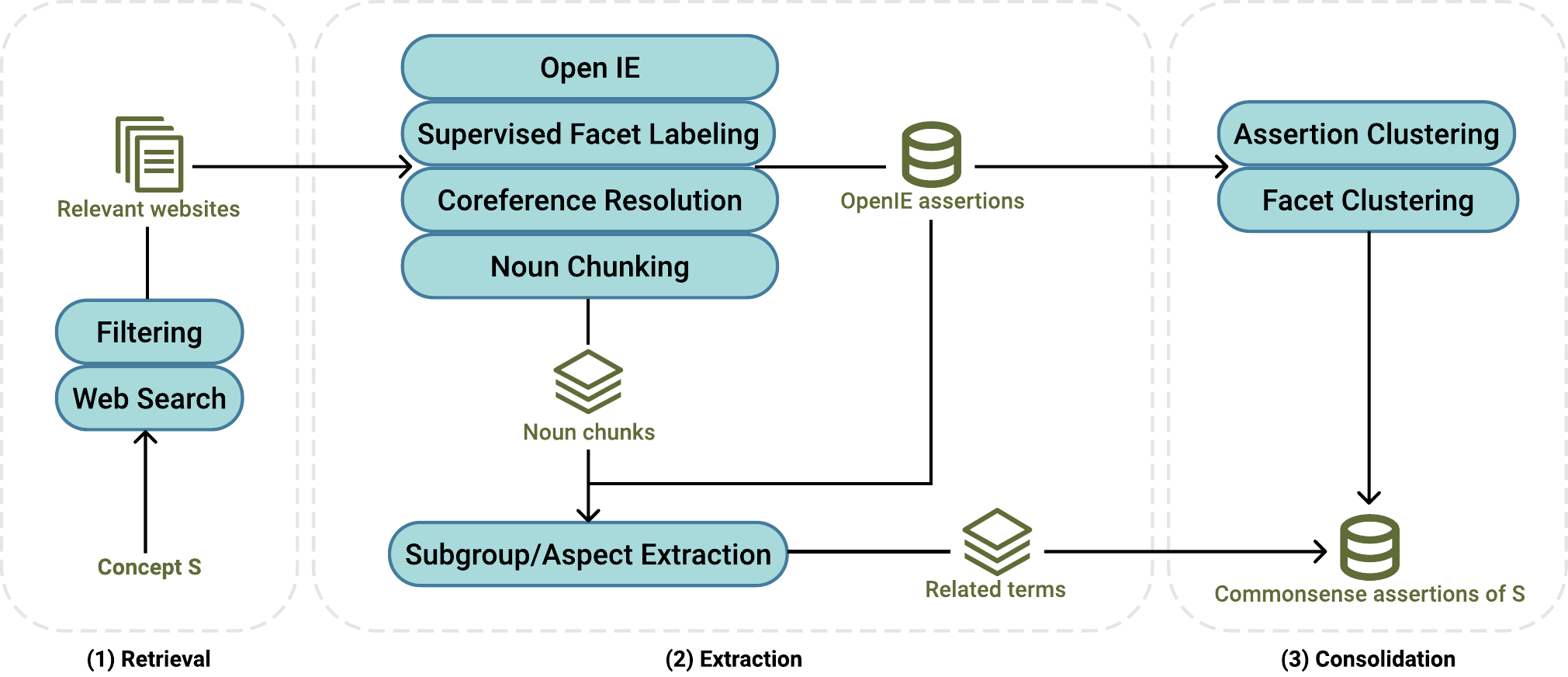}
    \caption{Architecture of our extraction pipeline.}
    \label{fig:architecture}
\end{figure*}

\paragraph{Approach}
The \ourkb{} method operates in three phases (illustrated in Fig.~\ref{fig:architecture}):
\squishlist
\item[1.] Source discovery:
    \squishlist
        \item[1a.] Retrieval of web 
        pages from search engines with specifically generated queries;
        \item[1b.] Filtering of result pages based on similarity to Wikipedia reference articles.
        \squishend
    \item[2.] Extraction of assertions with subgroups, aspects and facets:
    \squishlist
        \item[2a.] OIE for rule-based extraction using dependency-parsing patterns;
        \item[2b.] Labeling of semantic facets by supervised classifier.
    \squishend
    \item[3.] Clustering of assertions based on contextualized embeddings.
\squishend
%
The following section elaborates on these steps.

%% file: sections/4-Methodology.tex

\section{Methodology}

\subsection{Relevant Document Retrieval}
\paragraph{Web search} 
We use targeted web search to obtain documents specific to each subject, this way aiming to reduce the noise from out-of-context concept mentions, and the processing of large collections of mostly irrelevant documents, like encountered for instance in general web crawls. This is especially relevant as we later utilize coreference resolution, which is by itself a source of additional noise. Specifically, we utilize the Bing Web Search API.

Given a concept $s_0$, we first map it to a corresponding WordNet synset by taking the synset with the most lemma names, and use its hypernyms to refine search queries. For example, if $s_0$ has hypernym \textit{animal.n.01} then its search query is ``$s_0$ animal facts'', or if $s_0$ has hypernym \textit{professional.n.01} then its search query is ``$s_0$ job descriptions'', etc. We have manually designed templates for 35 commonly encountered hypernyms.
These cover 82.5\% of our subjects.
When none of the templates can be applied, we default to the direct hypernym of $s_0$ and form the following search query: ``$s_0$ (hypernym)''. Below we provide an example of search query for the animal lynx whose WordNet synset is \textit{lynx.n.02}, and a few top results returned by Bing.

\begin{quote}
    \texttt{\small \textbf{Query:} lynx animal facts \\
    \textbf{Top 5 results:}
    \begin{itemize}
        \item Lynx $\vert$ National Geographic
        \item Interesting facts about lynx $\vert$ Just Fun Facts
        \item Lynx Facts $\vert$ Softschools.com
        \item Facts About Bobcats \& Other Lynx $\vert$ Live Science
        \item Lynx $\vert$ Wikipedia
    \end{itemize}}
\end{quote}

\paragraph{Document filtering} Commercial search engines give us the benefit that (near-)duplicates, e.g., copies from Wikipedia, are well detected and ranked lower. At the same time, the search engine goal of diversification may introduce spurious results, despite our efforts with the search query refinement. This is exacerbated by our interest to obtain large sets of articles. We therefore propose a filter to remove irrelevant results. Given a subject $s_0$, we use the Bing API to retrieve 500 websites. For each website, we use a popular article scraping library\footnote{\href{https://github.com/codelucas/newspaper}{https://github.com/codelucas/newspaper}} to scrape its main content. Next, each retrieved document is compared with a Wikipedia reference article by the cosine similarity of the bag of words of both pages. 
As Wikipedia reference, we leverage the WordNet-Wikipedia pairings of BabelNet~\cite{navigli2012babelnet} and the resource by \cite{fernando2012mapping} as the first fallback. If both resources do not contain the desired WordNet synset, we simply pick the first Wikipedia article appearing in the search result.
After this, only documents with similarity higher than 0.55 (chosen based on tuning on withheld data) are retained.

\subsection{Knowledge Extraction}
\begin{table*}[ht!]
\centering
\footnotesize
\begin{tabular}{llll}
\toprule
\textbf{No.} & \textbf{Sentence} & \textbf{StuffIE~\cite{stuffie}} & \textbf{\ourkb{} OIE extractor} \\
\midrule
1 & They eat ptarmigans, voles, and grouse. & (1) They; eat; ptarmigans, voles, and grouse & \begin{tabular}[c]{@{}l@{}}(1) They; eat; ptarmigans\\ (2) They; eat; voles\\ (3) They; eat; grouse\end{tabular} \\
\midrule
2 & Lynx are active during evening and early morning. & \begin{tabular}[c]{@{}l@{}}(1) Lynx; are; active\\ (1.1) TEMPORAL: during evening and early morning\end{tabular} & \begin{tabular}[c]{@{}l@{}}(1) Lynx; are; active\\ (1.1) TEMPORAL: during evening\\ (1.2) TEMPORAL: during early morning\end{tabular} \\
\midrule
3 & Lions live for 20 years in captivity. & \begin{tabular}[c]{@{}l@{}}(1) Lions; live; \_\\ (1.1) PURPOSE: for 20 years\\ (1.2) LOCATION: in captivity\end{tabular} & \begin{tabular}[c]{@{}l@{}}(1) Lions; live; for 20 years\\ (1.1) LOCATION: in captivity\end{tabular} \\
\midrule
4 & Lions hunt many animals, such as gnus and antelopes. & (1) Lions; hunt; many animals, such as gnus and antelopes. & \begin{tabular}[c]{@{}l@{}}(1) Lions; hunt; gnus\\ (2) Lions; hunt; antelopes\end{tabular} \\
\midrule
5 & Dogs are extremely smart. & (1) Dogs; are; extremely smart & \begin{tabular}[c]{@{}l@{}}(1) Dogs; are; smart\\ (1.1) DEGREE: extremely\end{tabular} \\
\midrule
6 & Elephants are extremely good swimmers. & (1) Elephants; are; extremely good swimmers & \begin{tabular}[c]{@{}l@{}}(1) Elephants; are; good swimmers\\ (1.1) DEGREE: extremely\end{tabular}                                 \\
\bottomrule
\end{tabular}
\caption{Comparison of outputs returned by our OIE method and StuffIE.}
\label{tab:openie-examples}
\end{table*}

To enable the extraction of diverse pieces of information, our extraction step relies on open information extraction~\cite{DBLP:conf/ijcai/Mausam16,niklaus2018survey}. Similarly, as open assertions typically follow a general grammatical structure, we utilize dependency-path-based rules to identify extractions. We also rely on rules to identify aspects via possessive constructions, and subgroups via compound nouns. For assigning facets to semantic groups, we use supervised models, as the set of facets is small.

\paragraph{Rule-based statement extraction} Our open information extraction (OIE) method builts upon the StuffIE approach~\cite{stuffie}, a series of hand-crafted dependency-parse-based rules to extract triples and facets. The core ideas are to consider each verb as a candidate predicate of an assertion, and to identify subjects, objects and facets via grammatical relations, so-called dependency paths. The elaboration below uses the Clear style format (\url{http://www.clearnlp.com}), as used by the spaCy dependency parser:
\begin{itemize}
    \item Subjects are captured based on dependencies of the type subject (\code{nsubj}, \code{nsubjpass} and \code{csubj}) and  adjectival clauses (\code{acl}). If no subject is found, the parent verb of the predicate identified through adverbial clause modifier (\code{advcl}) and open clausal complement (\code{xcomp}) edges is used to identify subjects.
    \item Dependency edges used to find objects are direct object (\code{dobj}), indirect object (\code{iobj}), nominal modifier (\code{nmod}), clausal complement (\code{ccomp}) and adverbial clause modifier (\code{advcl}). 
    \item Once a triple has been formed, its constituents are completed by expanding their head words with related words via various dependency edges. For compound predicates, these include \code{xcomp}, \code{auxpass}, \code{mwe}, \code{advmod}. For compound subjects and objects, they are \code{compound}, \code{nummod}, \code{det}, \code{advmod}, \code{amod}.
    \item Finally, facets of a verb are identified through the following complements to the given verb: adverb modifier, prepositional and clausal complement.
\end{itemize}
%
We extend StuffIE's algorithm in the following ways: 
\begin{enumerate}
    \item The original algorithm includes all conjuncts of head words into one assertion, thus producing often overly specific assertions. In our method we break conjunctive objects (Table~\ref{tab:openie-examples}, row 1) and facets (Table~\ref{tab:openie-examples}, row 2) into separate assertions. Note that conjuncts should be connected by either \textit{``and''} or \textit{``or''}.
    \item The original algorithm frequently returns assertions with empty objects. To only return complete triples, in such cases, we identify the nearest prepositional facet after its predicate and convert the facet into the assertion's object (Table~\ref{tab:openie-examples}, row 3).
    \item We post-process special cases of sentences used for giving examples with the words: \textit{``like''}, \textit{``such as''} and \textit{``including''} to get finer-grained output (Table~\ref{tab:openie-examples}, row 4).
    \item We convert all adverb modifiers of objects (besides those of predicates as in StuffIE) into facets. There are two types of modifiers we consider:
    \begin{enumerate*}[label=(\roman*)]
        \item direct adverb modifiers connected to object's head word through the edge \code{advmod} (Table~\ref{tab:openie-examples}, row 5);
        \item the adverb in a noun phrase that follows the pattern ``adverb + adjective + object'' (Table~\ref{tab:openie-examples}, row 6).
    \end{enumerate*}
\end{enumerate}
%
%
Table~\ref{tab:openie-examples} gives a qualitative comparison of StuffIE's and our extraction results, while in the experiment section (Table~\ref{tab:openie}) we investigate their quantitative differences.

\paragraph{Subject and predicate postprocessing} After OIE, we perform coreference resolution\footnote{\href{https://huggingface.co/coref}{https://huggingface.co/coref}} on paragraph level to resolve nominative pronouns occuring as subjects. For instance, the primarily extracted assertion \triple{they, have, long trunks} will be replaced by \triple{the elephants, have, long trunks} if ``they'' is resolved to ``the elephants''. This step helps improve the number of assertions extracted for each concept.
Then, all subjects are normalized by removing determiners and punctuation, and by lemmatizing head nouns.  Moreover, predicates are normalized so that main verbs are transformed to their infinite forms (e.g., ``has been found in'' $\rightarrow$ ``be found in'', ``is performing'' $\rightarrow$ ``perform''). Finally all extracted facet words are removed from predicates and objects.

\paragraph{Facet type labeling} The extraction algorithm so far extracts facet values, but is unaware of their semantic type (e.g., ``spatial'' or ``causal''). For assigning semantic types, we fine-tune a RoBERTa~\cite{liu2019roberta} model to classify each facet into one of the aforementioned eight types. The input sequences of RoBERTa take the form: ``[CLS] subject [PRED] predicate [OBJ] object [FCT] facet [SEP]'', where [PRED], [OBJ] and [FCT] are special tokens used for marking the borders between different elements. The output vector of the [CLS] token is then passed to a fully-connected layer stacked with a soft-max layer on top of the transformer architecture to label the facet. Details on classifier training are in Section~\ref{sec:per-module-eval}.

\paragraph{Extraction of subgroups} Subgroups could be sub-species in case of animals, or refer to the target concept in different states, such as ``hunting cheetah'' and ``retired policeman''. For subject $s_0$, we collect all noun chunks (normalized as for triple subjects described above) ending with $s_0$ or any of its WordNet lemmas as potential candidates. 
Semantically similar chunks, such as ``Canadian lynx'' and ``Canada lynx'', are then grouped using hierarchical agglomerative clustering (HAC) on average word2vec representations.
In addition, we leverage WordNet to distinguish antonyms, with which vector space embeddings typically struggle. Note that the subgroups are restricted to be less-than-5-words, and subgroups that syntactically contain other subgroups are disregarded (e.g., ``old male Canadian lynx'' is grouped with ``Canadian lynx''). In addition, a chunk will be ignored if it is a named entity (e.g., ``Will Smith'' for the concept ``smith''). Finally we use WordNet hyponyms to remove spurious subgroups, e.g., ``sea lion'' and ``ant lion'' w.r.t.\ ``lion''.

\paragraph{Extraction of related aspects} Given subject $s_0$ and its WordNet lemmas $L_{s_0}$, related aspects of the subject are extracted from noun chunks collected from two sources:
\begin{enumerate}[label=(\roman*)]
    \item \textit{Possessive noun chunks} where the possessives refer to any lemma in $L_{s_0}$, for example, ``elephant's diet'' and ``their diet'' (with resolution to ``elephant'');
    \item \triple{$s$, $p$, noun chunk} triples where $s \in L_{s_0}$ and $p$ is one of the following verb phrases: ``have'', ``contain'', ``be assembled of'' or ``be composed of''.
\end{enumerate}
In order to prevent too specific aspects (e.g., ``large paws'' or ``short tails''), only compound nouns (if applicable) or nouns in these \textit{noun chunks} are then extracted as aspects of $s_0$. For example, if we observe \triple{lynx, have, black ear tuft}, then the adjective ``black'' is ignored and ``ear tuft'' will be extracted instead of only extracting the head noun ``tuft''.

\paragraph{Retained assertions}
For each primary subject, a separate set of documents is processed, and the output of this stage are three sets of assertions: Assertions for the primary subject $s_0$, assertions for its subgroups, and assertions for its aspects. These are selected as follows.


As assertions for the main subject and its subgroups we simply retain all assertions that have a subject that matches a WordNet lemma of the primary subject, or the name of one of its subgroups.


The case of aspect assertions is slightly more complex, we merge three cases:
\begin{enumerate}
    \item Assertions that have a subject which is among the previously identified aspects;
    \item Assertions that have a subject among the lemmas of the main subject, and an object which is a noun chunk consisting of an aspect $t \in asp_{s_0}$ as the head noun and an adjectival modifier $adj$ of $t$. For instance, from the assertion \triple{elephant, have, a long very trunk} we infer that \triple{elephant trunk, be, long, DEGREE: very}.
    \item All noun chunks that follow the pattern \emph{``possessive + adj + t''} (e.g., ``elephant's long trunks''), where $possessive$ refers to any lemma in $L_{s_0}$, \textit{adj} is an adjectival modifier of $t$, and $t \in asp_{s_0}$.
\end{enumerate}
Results from the latter two cases are transformed into \triple{t, be, adj, F} assertions where the facets $F$ are extracted from adverb modifiers of \textit{adj}.

\subsection{Knowledge Consolidation}
Natural language is rich in paraphrases, and consequently, the extraction pipeline so far produces frequently assertions that carry the same or nearly the same meaning. Identifying and clustering such assertions is necessary, in order to avoid redundancies, and get better frequency signals for individual assertions.

\paragraph{Triple clustering} Because extraction is done for each concept separately, we only need to cluster predicate-object pairs. First, we train a RoBERTa model to detect if two given triples are semantically similar (for setup details see Sec.~\ref{sec:per-module-eval}). Confidence scores given by the model are then used to compute distances for the HAC algorithm to group assertions into clusters. Given two assertions \triple{$s, p_1, o_1$} and \triple{$s, p_2, o_2$}, the input sentence given to RoBERTa is: ``[CLS] [SUBJ] $p_1$ [U-SEP] $o_1$ [SEP] [SUBJ] $p_2$ [U-SEP] $o_2$ [SEP]'', where [SUBJ] and [U-SEP] are new special tokens introduced to replace identical subjects and mark the borders between predicates and objects, respectively. The output vector of the [CLS] token is used for the classification purpose in the same way as in the model used for facet labeling described above.

Ideally one would compute the full distance matrix between all assertions (an $n\times n$ matrix for $n$ triples), but given that pretrained language models (LM) are exceedingly resource-intensive, this quadratic computation would be expensive even for moderate assertion sets. We therefore reduce the computational effort by pre-filtering the set of pairs to be compared by the pretrained LM.
\begin{enumerate}
    \item The assertions are sorted in decreasing order of frequency.
    \item We compute cosine similarities between vector representations of predicate-object pairs, using word2vec embeddings. This can be done very fast with parallel matrix multiplication.
    \item For each assertion $a_i$, we then only compute the RoBERTa-based distances with the top-$k$ most similar assertions (ranked by word2vec-based similarities) that succeed $a_i$ in the sorted list (the sorted list helps us focus on salient assertions). All other pairs get the distance of $1.0$. This produces a ``sparse'' distance matrix for $n$ assertions.
    \item For clustering, we use the HAC algorithm with single linkage, because it only looks at the most similar pairs between two clusters. That helps to reduce the chance of missing similar triples whose similarities were not computed by RoBERTa in the third step.
\end{enumerate}
After clustering, the most frequent assertion inside each cluster is used as representative.

\paragraph{Facet value clustering}
Facet values may similarly exhibit redundancy, for example, the degree facet may come with values ``often'', ``frequently'', ``mostly'', ``regularly'', etc. Also, sources may occasionally mention odd values. We combat both by clustering facet values per facet type, and retaining only the one with strongest support.

Considering the small number of facet values per assertion and facet type (usually less than 5), we utilize simple methods for clustering. Specifically, given the list of values, we use the HAC algorithm to cluster values which are adverbs, in which distance between two values is measured by the cosine distance of their word2vec presentations. Other values are grouped if they have the same head word (e.g., ``during evening'' and ``in the evening'' go to one same cluster). Similarly, the most frequent value inside a cluster is used as representative of that cluster. 

%% file: sections/5-Experiments.tex
\section{Experiments}

The evaluation of \ourkb{} is centered on three research questions:
\begin{itemize}
    \item \textbf{RQ1:} Is the resulting CSKB of higher quality than existing resources?
    \item \textbf{RQ2:} Does (structured) CSK help in extrinsic use cases?
    \item \textbf{RQ3:} What is the quality and extrinsic value of  facets?
\end{itemize}
We first present the implementation of \ourkb{}, then discuss each of these research questions in its own subsection.

\subsection{Implementation}

We executed the pipeline for the 10,000 most popular subjects in ConceptNet (ranked by number of assertions). The execution of took a total of 10 days, of which about 5 days were spent on website crawling, 3 days on statement extraction, and 2 days on clustering. For each subject, we used the Bing Search API to retrieve 500 websites. The resulting CSKB contains 3,693,990 assertions for these primary subjects, and 1,768,538 assertions for 280,970 subgroups and 3,349,198 for 92,038 aspects. On average, half of all assertions have a facet (see Table~\ref{tab:descent-statistics}). 

\begin{table}[t]
\centering
\small
    \begin{tabular}{lrrr}
    \toprule
    \textbf{Subject type} & \textbf{\#s} & \textbf{\#spo} & \textbf{\#facets} \\
    \midrule
    Primary & 10,000 & 3,693,990 & 2,169,119 \\
    Subgroup & 280,970 & 1,768,538 & 944,124 \\
    Aspect & 92,038 & 3,349,198 & 1,467,159 \\
    All & 382,555 & 8,562,593 & 4,425,628 \\
    \bottomrule
    \end{tabular}
    \caption{Statistics of \ourkb{} KB.}
    \label{tab:descent-statistics}
\end{table}

In Table~\ref{tab:relativesizecomparison}, we show statistics of our CSKB in comparison with popular existing resources. For comparability, we report statistics on a sample of 50 popular animals and 50 popular occupations introduced in \cite{quasimodo}, in addition to 50 popular concepts in the engineering domain collected using Wiktionary word frequencies (e.g., car, bus, computer, phone, etc.). 
For statistics, subgroups are collected through \textit{hyponyms} (WordNet) and relation \textit{IsA} (ConceptNet and TupleKB). Aspects are collected via \textit{part meronyms} (WordNet), relation \textit{PartOf} (ConceptNet), \textit{hasPart} (TupleKB), \textit{hasPhysicalPart} (WebChild) and \textit{hasBodyPart} (Quasimodo). We divide the statistics of our KB into three categories: general assertions (\ourkb), subgroup assertions (\ourkb$^{sg}$) and aspect assertions (\ourkb$^{asp}$).
Table~\ref{tab:relativesizecomparison} shows 
that \ourkb{}, among all resources, is the only one which conveys qualitative facets besides triples. \ourkb{} also extracts a considerable amount of assertions for the primary subjects. In addition, \ourkb{} has the capability to extend the 150 primary subjects to 13,869 subgroups and related aspects, approximately tripling the number of the extracted assertions. We extract more subgroups than any other KB. Regarding aspects we are only outperformed by WebChild, which includes many uninformative and rather ``exotic'' part-of triples (e.g., teacher has cell, lion has facial vein).

\begin{table}[t]
\centering
\small
    \begin{tabular}{lrrrrr}
    \toprule
    \textbf{Resource} & \textbf{\#s} & \textbf{\#spo} & \textbf{\#facets} & \textbf{\!\!\!\#subgroups} & \textbf{\!\!\#aspects} \\
    \midrule
    WordNet~\cite{wordnet} & 150 & - & - & 1,472 & 229 \\
    WebChild~\cite{webchild} & 150 & 178,073 & - & - & 47,171 \\
    ConceptNet~\cite{conceptnet} & 150 & 7,313 & - & 7,239 & 368 \\
    TupleKB~\cite{mishra2017domain} & 133 & 23,106 & - & 231 & 2,302 \\
    Quasimodo~\cite{quasimodo} & 150 & 137,880 & - & - & 563 \\
    GenericsKB~\cite{bhakthavatsalam2020genericskb} & 150 & 192,075 & - & - & - \\
    \textbf{\ourkb} & 150 & 132,070 & 80,717 & 10,026 & 5,843 \\
    \hdashline
    \textbf{\ourkb$^{sg}$} & 8,251 & 110,631 & 64,449 & - & - \\
    \textbf{\ourkb$^{asp}$} & 5,618 & 169,770 & 74,449 & - & - \\
    \bottomrule
    \end{tabular}
    \caption{Statistics of different resources on top 50 subjects for 
    three domains: animals, occupations, engineering.} 
    \label{tab:relativesizecomparison}
\end{table}

\subsection{Intrinsic Evaluation}

To investigate RQ1, we instantiate \emph{quality} with the standard notions of precision and recall, splitting precision further up into the dimensions of \emph{typicality} and \emph{salience}, measuring this way the degree of truth, and the degree of relevance of assertions (cf.~\cite{quasimodo}). 
Typicality states that an assertion holds for most instances of a concept. For example, elephants using their trunk is typical, whereas elephants drinking milk holds only for baby elephants.
Salience refers to the human perspective of whether an assertion is associated with a concept by most humans more or less on first thought.
For example, elephants having trunks is salient, whereas 
elephants killing their mahouts (trainers) is not.

\begin{figure}[t]
\scriptsize
\begin{subfigure}{\columnwidth}
\centering
\begin{tikzpicture}
\begin{axis}[
    ybar,
    bar width=0.25cm,
    height=3.3cm,
    width=\columnwidth,
    enlarge x limits=0.6,
    ylabel={quality@5},
    symbolic x coords={typicality,salience}, 
    ylabel near ticks, 
    xtick=data,
    ]
\addplot[black,fill=violet,pattern=north east lines]
    coordinates {
        (typicality, 4.053333333333334)
        (salience, 3.6911111111111112)
    }; \label{conceptnet-bar}
\addplot[olive,fill=olive] 
    coordinates {
        (typicality, 3.077777777777778)
        (salience, 2.8422222222222224)
    }; \label{webchild-bar}
\addplot[teal,fill=teal] 
    coordinates {
        (typicality, 3.688888888888889)
        (salience, 2.6733333333333333)
    }; \label{tuplekb-bar}
\addplot[blue,fill=blue] 
    coordinates {
        (typicality, 3.18)
        (salience, 2.92)
    }; \label{qmodo-bar}
\addplot[red,fill=red]  
    coordinates {
        (typicality, 3.513333333333333)
        (salience, 3.3066666666666666)
    }; \label{descent-bar}
\addplot[magenta,fill=magenta]  
    coordinates {
        (typicality, 3.6466666666666665)
        (salience, 3.3088888888888888)
    }; \label{descent-sg-bar}
\addplot[violet,fill=violet]
    coordinates {
        (typicality, 3.3155555555555556)
        (salience, 2.98)
    }; \label{descent-sp-bar}
\end{axis}
\end{tikzpicture}
\caption{Precision evaluation for triples} 
\label{fig:eval-correctness-comp}
\end{subfigure}

\begin{subfigure}{\columnwidth}
\centering
\begin{tikzpicture}
\begin{axis}[
    ybar,
    bar width=0.15cm,
    height=3.3cm,
    width=.55\columnwidth,
    enlarge x limits=0.5,
    ylabel={recall (\%)},
    ylabel near ticks,
    symbolic x coords={strict,relaxed},
    xtick=data,
    ]
\addplot[black,fill=violet,pattern=north east lines]
    coordinates {
        (strict, 11.50)
        (relaxed, 24.92)
    };
\addplot[olive,fill=olive] 
    coordinates {
        (strict, 12.73)
        (relaxed, 19.09)
    };
\addplot[teal,fill=teal] 
    coordinates {
        (strict, 19.16)
        (relaxed, 33.49)
    };
\addplot[blue,fill=blue] 
    coordinates {
        (strict, 26.63)
        (relaxed, 55.25)
    };
\addplot[red,fill=red]  
    coordinates {
        (strict, 29.50)
        (relaxed, 72.06)
    };
\end{axis}
\end{tikzpicture}
\quad
\begin{tikzpicture}
\begin{axis}[
    ybar,
    bar width=0.15cm,
    height=3.3cm,
    width=.55\columnwidth,
    enlarge x limits=0.5,
    symbolic x coords={strict$@$10,relaxed$@$10},
    xtick=data,
    ]
\addplot[black,fill=violet,pattern=north east lines]
    coordinates {
        (strict$@$10, 8.33)
        (relaxed$@$10, 16.09)
    };
\addplot[olive,fill=olive] 
    coordinates {
        (strict$@$10, 6.03)
        (relaxed$@$10, 6.06)
    };
\addplot[teal,fill=teal] 
    coordinates {
        (strict$@$10, 8.38)
        (relaxed$@$10, 8.99)
    };
\addplot[blue,fill=blue] 
    coordinates {
        (strict$@$10, 9.51)
        (relaxed$@$10, 13.56)
    };
\addplot[red,fill=red]  
    coordinates {
        (strict$@$10, 10.26)
        (relaxed$@$10, 15.62)
    };
\end{axis}
\end{tikzpicture}
\caption{Recall evaluation for triples}
\label{fig:eval-recall}
\end{subfigure}
\begin{subfigure}{\columnwidth}
\centering
    \caption*{
        \scriptsize
        \ref{conceptnet-bar} ConceptNet
        \enspace
        \ref{webchild-bar} WebChild
        \enspace
        \ref{tuplekb-bar} TupleKB
        \enspace
        \ref{qmodo-bar} Quasimodo
    }
    \caption*{
        \scriptsize
        \ref{descent-bar} \ourkb
        \enspace
        \ref{descent-sg-bar} \ourkb$^{sg}$
        \enspace
        \ref{descent-sp-bar} \ourkb$^{asp}$
    }
\end{subfigure}
\caption{Precision and recall assessment of different CSKBs.}
\label{fig:kb-eval}
\end{figure}
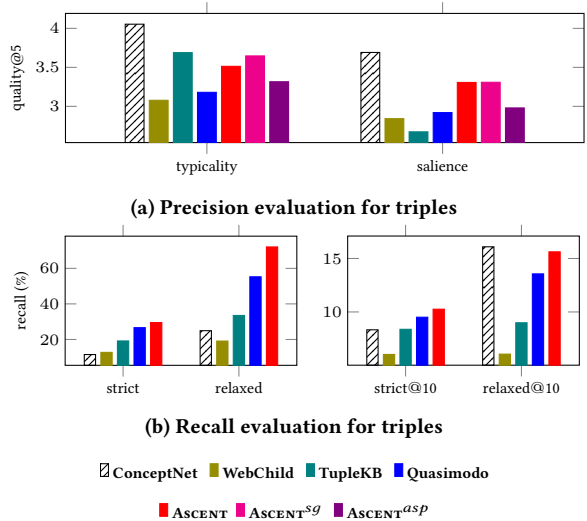

\paragraph{Assertion precision} 
Unlike for encyclopedic knowledge (``The Lion King'' was either produced by Disney, or it wasn't), precision of CSK is generally not a binary concept, calling for more refined evaluation metrics. We follow  the Quasimodo project \cite{quasimodo} which assessed typicality and salience. Given a CSK triple, annotators on Amazon MTurk are asked to evaluate each of the two aspects on a scale from 1 (lowest) to 5 (highest). We use the same sampling setup as proposed in \cite{quasimodo}: for each KB (i.e., \ourkb{} and the prior CSKBs), create a pool that contains the 5 top-ranked triples of each of a selected set of subjects, then randomly sample 50 triples from this pool. In addition, specifically for our KB, we create a pool from top-5 ranked subgroup assertions of each subject, then also draw 50 random triples from the pool for evaluation, which is reported as \ourkb$^{sg}$. The same sampling process is applied for aspect assertions in our KB, which is reported as \ourkb$^{asp}$. Each triple is evaluated by three different crowd-workers. We iteratively evaluate triple quality for three sets of 50 subjects of three domains: animals, occupations and engineering, respectively. We report the aggregated results in Fig.~\ref{fig:eval-correctness-comp}. Among the automatically-constructed KBs (i.e., except for ConceptNet), our KB has the most salient assertions while demonstrating competitive quality when it comes to typicality. These results indicate that our source selection, filtering and extraction scheme allows to pull out important assertions better than other CSKBs.

\paragraph{Assertion recall} Evaluating recall requires a notion of ground truth. For this purpose, we use crowdsourcing-based phrases from humans collected by Quasimodo~\cite{quasimodo}:
2,400 free association sentences for 50 occupations and 50 animals. We also evaluate using the same metrics, strict and relaxed sentence-assertion match. In the \textit{relaxed} mode, we measure the fraction of tokens, from the human-written phrase, that are contained in some KB triples for the corresponding subject.
In the \textit{strict} mode, we only consider statements where P, O or PO is exactly found in the human-written phrase, and measure the fraction of matching characters vs.\ the total length of the human-written phrase. 
To match natural language with KB predicates, we use generic translations  (e.g., \textit{hasProperty} $\rightarrow$ \textit{is}, \textit{hasPhysicalPart} $\rightarrow$ \textit{has}, \textit{is-part-of} $\rightarrow$ \textit{is part of}, etc.). The evaluation results can be seen in Fig.~\ref{fig:eval-recall}. We observe that \ourkb{} captures a significantly higher fraction of the ground-truth assertions provided by crowd workers than any of the other CSKBs. When we limit CSKBs to their top-10 ranked triples for each subject, \ourkb{} outperforms all other KBs in the strict mode and is the second-best after ConceptNet, which is the only one that was constructed manually, in the relaxed mode. This result affirms that our top-ranked assertions 
have high quality compared to other CSKBs.

\paragraph{Subgroups and aspects}
We compare \ourkb{} \textit{subgroup} entries to the manually created ConceptNet, and against a comprehensive taxonomy, WebIsALOD~\cite{HertlingPaulheim:ISWC2017}, 
automatically built by applying 58 Hearst-style extraction patterns to the Common Crawl corpus. For a random sample of 500 subgroup entries per resource, we manually found an average precision of 5.6\% for WebIsALOD, 83.4\% for ConceptNet, and 92.0\% for \ourkb{} (note that we manually filtered out instances in WebIsALOD's entries). Our precision significantly outperforms WebIsALOD, and is even better than the manually constructed ConceptNet. At the same time, it is worth to point out that our approach misses out on subgroups that do not lexically contain the main subject, e.g., ``panda'' as subgroup of ``bear''.


We compare \textit{aspects} against two resources: hasPartKB~\cite{haspartkb} and predictions made by masked 
language models (LMs). As neural-embedding LM, we use RoBERTa-Large and follow the idea of \cite{weir2020probing} to ask the LM to predict the missing word in the sentence ``Everyone knows that <subject> has 
<?>.''
We use the human-generated CSLB concept property norm dataset \cite{devereux2014centre} as ground truth, retaining only headwords to allow a fair comparison with the masked prediction that produces only a single token. 
 Since \ourkb{} contains a wider range of aspects than just physical parts as in hasPartKB and the CSLB dataset, we use recall@$k$ as the metrics for this evaluation, focusing on the top-5 terms from CSLB. 
%
 Considering the top-5, top-10 and top-20 assertions per KB/LM, \ourkb{} achieves recall@5 of 0.27, 0.41, 0.53, compared with hasPartKB at 0.13, 0.22, 0.35, and RoBERTa-Large at 0.29, 0.41, 0.51. Thus, \ourkb{} considerably outperforms hasPartKB in this setup, and performs on par with state-of-the-art language models.



\subsection{Extrinsic Evaluation}
\label{subsec:extrinsic}

To answer RQ2, we conduct a comprehensive evaluation of the contribution of commonsense knowledge to question answering (QA) via four different setups, all based on the idea of  priming pre-trained LMs with context~\cite{guu2020realm,petroni2020context}:
\begin{enumerate}
    \item In \textit{masked prediction} (MP) \cite{petroni2019language}, we ask language models to predict single tokens in generic sentences.
    \item In \textit{free generation} (FG), we provide only questions, and let LMs generate arbitrary answer sentences.
    \item In \textit{guided generation} (GG), LMs are provided with an answer sentence prefix. This provides a middle ground between the previous two setups, allowing multi-token answers, but avoiding some overly evasive answers.
    \item In \textit{span prediction} (SP), LMs select best answers from provided content~\cite{lan2019albert}.
\end{enumerate}
We illustrate all settings in Table~\ref{tab:qa-settings}. In all settings, LMs are provided with context in the form of assertions taken from either ConceptNet, TupleKB, Quasimodo, GenericsKB or \ourkb{}. 
These setups are motivated by the observation that priming language models with context can significantly influence their predictions~\cite{guu2020realm,petroni2020context}. Previous works on language model priming mostly focused on evaluating retrieval strategies. In contrast, our comprehensive test suite focuses on the impact of utilizing different CSK resources, while leaving the retrieval component constant.

Masked prediction is perhaps the best researched problem, coming with the advantage of allowing automated evaluation, although automated evaluation may unfairly discount sensible alternative answers. Also, masked prediction is limited to single tokens. Free generations circumvent this restriction, although they necessitate human annotations, and are prone to evasive answers. They are thus well complemented by extractive answering schemes, which limit the language models abstraction abilities, but provide the cleanest way to evaluate the context alone.

\begin{table}[t]
\centering
\scriptsize
\begin{tabular}{cp{.5\columnwidth}p{.3\columnwidth}}
\toprule
\textbf{Setting} & \textbf{Input} & \textbf{Sample output} \\
\midrule
MP & Elephants eat {[}MASK{]}. {[}SEP{]} Elephants eat roots, grasses, fruit, and bark, and they eat a lot of these things. & everything (15.52\%), trees (15.32\%), plants (11.26\%) \\
\hline
\multirow{4}{*}{FG} & C: Elephants eat roots, grasses, fruit, and bark, and they eat a lot of these things. & They eat a lot of grasses, fruits, and trees. \\
 & Q: What do elephants eat? & \\
 & A: & \\
\hline
\multirow{4}{*}{GG} & C: Elephants eat roots, grasses, fruit, and bark, and they eat a lot of these things. & Elephants eat a lot of things. \\
 & Q: What do elephants eat? & \\
 & A: Elephants eat & \\
\hline
\multirow{3}{*}{SP} & question=``What do elephants eat?'' & start=14, end=46, \\
 & context=``Elephants eat roots, grasses, fruit, and bark, and they eat a lot of these things.'' & answer=``roots, grasses, fruit, and bark'' \\
\bottomrule
\end{tabular}
\caption{Examples of 4 QA settings (MP - masked prediction, FG - free generation, GG - guided generation, SP - span prediction). Sample output was given by RoBERTa (for MP), GPT-2 (for FG and GG) and ALBERT (for SP).}
\label{tab:qa-settings}
\end{table}

\paragraph{Models} Following standard usage, we use RoBERTa-Large for masked prediction, the autoregressive GPT-2 for the two generative setups, and ALBERT-xxlarge~\cite{lan2019albert}, fine-tuned on SQuAD 2.0 for span prediction.

\paragraph{Context retrieval method}
Given a query, we use a simple token overlapping method to pull out relevant assertions from a CSKB. First, we only take into account assertions whose subjects are mentioned in the query. We rank these assertions by the number of distinct tokens occurring in the input query (ignoring stop words). For each query, we pick up the top ranked assertions and concatenate them to build the context. For comparability, we limit the length of every context to 256 characters. As rank tie-breaker, we use original ranks in the CSKBs.

\paragraph{Task construction}
Previous work has generated masked sentences based on templates from ConceptNet triples~\cite{petroni2019language}. However, the resulting sentences are often unnatural, following the idiosyncrasies of the ConceptNet data model. We therefore built a new dataset of natural commonsense sentences for \textit{masked prediction}. We use the CSLB property norm dataset~\cite{devereux2014centre} which consists of short human-written sentences about salient properties of general concepts. We hide the last token of each sentence, which is usually the object of that sentence. Besides, we remove sentences that contain less than three words. The resulting dataset consists of 19,649 masked sentences.

For the \textit{generative and extractive settings}, we use the Google Search Auto-completion functionality to collect commonsense questions about the aforementioned set of 150 engineering concepts, animals and occupations. For each subject, we feed the API with 6 prefixes: ``what/when/where are/do <subject>'', then we collect all auto-completed queries returned by the API.
We got 8,098 auto-completed queries for these subjects. Next, we drew samples from that query set, then manually removed jokes and other noise (e.g., ``where do cows go for entertainment'')
obtaining
50 questions for evaluation. The answers from each KB in each generative or extractive setting were then posted on Amazon MTurk, along with test questions that ensured answer quality.

\paragraph{Evaluation scheme}
For commonsense topics, questions often have multiple valid answers. Additionally, given that answers in our settings of generative and extractive QA are very open, creating an automated evaluation is difficult. We therefore use human judgements for evaluating all settings except masked prediction. Specifically, given a question and set of answers, we ask humans to assess each answer based on two dimensions, \textit{correctness} and \textit{informativeness}, on a scale from 1 (lowest) to 5 (highest). Each question is evaluated by three annotators in Amazon MTurk. For evaluating masked prediction, we use the mean precision at k ($P@k$) metric, following \cite{petroni2019language}.

\paragraph{Results}
The evaluation results are shown in Table~\ref{tab:qa-results}. We can see that all KBs contribute contexts that improve LM response quality. \ourkb{} performs significantly better than the no-context baseline in both FG, GG and MP settings (p-values of paired t-test below 0.013), Besides, in the span prediction (SP) setting, where answers come directly from retrieved contexts, \ourkb{} outperforms all competitors, indicating that our assertions have very high quality compared to other KBs -- with statistically significant gains (p-value below 0.038) over TupleKB on both metrics, and Quasimodo on correctness. Notably, our structured resource also outperforms the text-based GenericsKB in all but one case.
For the MTurk assessments, we obtained a mean score variance of 0.76 and a mean Pearson correlation coefficient of 0.58, which indicate high agreement among annotators. We demonstrate three examples for the retrieved contexts and answers generated by GPT-2 in Table~\ref{tab:qa-examples}.


\begin{table}[t]
    \centering
    \small
    \begin{tabular}{lccccccc}
    \toprule
    \multicolumn{1}{l}{\multirow{2}{*}{\textbf{Context}}} & \multicolumn{2}{c}{\textbf{FG}} & \multicolumn{2}{c}{\textbf{GG}} & \multicolumn{2}{c}{\textbf{SP}} & \multicolumn{1}{c}{\textbf{MP}} \\
    \cline{2-8}
     & \textbf{C} & \textbf{I} &  \textbf{C} & \textbf{I} &  \textbf{C} & \textbf{I} & \textbf{P@5} \\
    \midrule
    No context & 2.44 & 2.22 & 2.87 & 2.57 & - & - & 17.9 \\
    ConceptNet & 2.74 & 2.39 & 3.03 & 2.61 & 2.34 & 2.16 & 24.5 \\
    TupleKB & 2.84 & 2.53 & \textbf{3.46} & \textbf{3.03} & 1.82 & 1.62 & 23.7 \\
    Quasimodo & 2.58 & 2.31 & 3.06 & 2.72 & 2.22 & 2.05 & 25.1 \\
    GenericsKB-Best & 2.89 & \textbf{2.71} & 3.13 & 2.77 & 2.39 & 2.20 & 24.8 \\
    \textbf{\ourkb}$^{\textit{tri}}$ & \textbf{2.91} & 2.68 & 3.41 & 3.01 & \textbf{2.61} & \textbf{2.34} & \textbf{25.9} \\
    \bottomrule
    \end{tabular}
    \caption{Results of our QA evaluation. Metrics: C - correctness, I - informativeness, P@5 - precision at five (\%). \ourkb$^{tri}$ contains only triples in \ourkb{}.} 
    \label{tab:qa-results}
\end{table}


\begin{table}[t]
\centering
\scriptsize
\begin{tabular}{lp{.48\columnwidth}p{.35\columnwidth}}
\toprule
 & \textbf{Question + Retrieved contexts} & \textbf{Answer} \\
\toprule \toprule
 & \textsc{\textbf{When are rats awake?}} & \\
 \hline
\multirow{1}{*}{\rotatebox{90}{\tiny $\emptyset$}} & \textit{-} & The rats are awake. \\
\hline
\multirow{3}{*}{\rotatebox{90}{\tiny CN}} & Awakes have yawn. Awakes need open eyes. Awakes need sleep. Awaking has rub eyes. Awakes need set alarm clock. Ratss are rodents... & Rats are awake. \\
\hline
\multirow{3}{*}{\rotatebox{90}{\tiny TP}} & Rats have bone. Rats have brain. Rats have cell. Rats have chest cavity. Rats have chest. Rats have cytoplasm. Rats have ear. & Rats are awake. \\
\hline
\multirow{3}{*}{\rotatebox{90}{\tiny QS}} & Rats carry rabies. Rats can jump. Rats get electrocuted. Rats get rabies. Rats are pet. Rats have gallbladder. Rats have bladder... & Rats are awake. \\
\hline
\multirow{3}{*}{\rotatebox{90}{\tiny GB}} & Rats are omnivorous, which means they eat both plant and animal material. Rats are nocturnal animals. Rats are monophyodont, whi... & Rats are awake at night. \\
\hline
\multirow{3}{*}{\rotatebox{90}{\tiny AT}} & Rats are intelligent. Rats are clean. Rats are nocturnal. Rats are active. Rats are careful. Rats grind their teeth. Rats are om... & Rats are awake.  \\
\hline
\multirow{3}{*}{\rotatebox{90}{\tiny AQ}} & Rats are very intelligent. Rats are very clean. Rats are mainly nocturnal. Rats are active at night. Rats are very careful. Rats... & Rats are awake at night. \\
\toprule \toprule
 & \textsc{\textbf{Where do shepherds live?}} & \\
 \hline
\multirow{1}{*}{\rotatebox{90}{\tiny $\emptyset$}} & \textit{-} & They live in the woods. \\
\hline
\multirow{3}{*}{\rotatebox{90}{\tiny CN}} & Lives have die. Lives have breathe. Lives need born. Living causes dying. Living causes death. Living causes life. Lives have bo...  & They live in the wilds of the North. \\
\hline
\multirow{3}{*}{\rotatebox{90}{\tiny TP}} & Shepherds have arm. Shepherds have body substance. Shepherds have cell membrane. Shepherds have cell. Shepherds have corpus. She... & Shepherds live in the wild. \\
\hline
\multirow{3}{*}{\rotatebox{90}{\tiny QS}} & Shepherds can live. Noes can live. Living is thing. Lives are illegal in uk. Living is alone. Living is abroad. Living is good. & Shepherds live in the wild. \\
\hline
\multirow{3}{*}{\rotatebox{90}{\tiny GB}} & Shepherds care for their sheep but also expect to make a living. Shepherds require proteins, carbohydrates, minerals, vitamins a...  & Shepherds live in the wild, but they are not allowed to breed in the wild. \\
\hline
\multirow{3}{*}{\rotatebox{90}{\tiny AT}} & Shepherds live in crop fields. Shepherds care for the sheep. Shepherds are the flock of god. Being is living. Shepherds keep the...  & Shepherds live in crop fields. \\
\hline
\multirow{3}{*}{\rotatebox{90}{\tiny AQ}} & Shepherds live in crop fields. Shepherds care for the sheep to green pastures. Shepherds are the flock of god. Being is living f...  & Shepherds live in crop fields. \\
\toprule \toprule
 & \textsc{\textbf{What are cars used for?}} & \\
 \hline
\multirow{3}{*}{\rotatebox{90}{\tiny $\emptyset$}} & \textit{-} & The most common car used for driving is a car with a front end that is not a front end. \\
\hline
\multirow{3}{*}{\rotatebox{90}{\tiny CN}} & Cars are used for fun. Cars are used for getting to work. Carss can use fuel. Cars are used for drive. Cars are used for transpo... & Cars are used for fun. \\
\hline
\multirow{3}{*}{\rotatebox{90}{\tiny TP}} & Cars use chemical energy. Cars use energy. Cars use gasoline. Cars use suspension. Cars use independent suspension. Cars use pis... & Cars used for:  1. A car that is used for:  1. A car that is used for: ...\\
\hline
\multirow{3}{*}{\rotatebox{90}{\tiny QS}} & Cars use gas. Cars use diesel. Cars use fuel. Cars use circuit breakers. Cars use transmission. Cars use engines. Cars use propa... & Cars used for the purpose of transporting goods. \\
\hline
\multirow{3}{*}{\rotatebox{90}{\tiny GB}} & Cars are complex devices used in a complex environment. Cars use up energy and cause pollution. Cars use up energy and cause pol... & Cars are complex devices used in a complex environment. \\
\hline
\multirow{3}{*}{\rotatebox{90}{\tiny AT}} & Cars are used for transportation. Cars use parallel washington street. Cars use part. Cars use lever. Cars use less manpower. Ts... & Cars are used for transportation. \\
\hline
\multirow{3}{*}{\rotatebox{90}{\tiny AQ}} & Cars are used for transportation. Cars use parallel washington street to a crest at hyde street. Cars use part by drew winter an... & Cars are used for transportation. \\
\bottomrule
\end{tabular}
\caption{Examples of retrieved KB assertions and answers generated by GPT-2. Abbreviations: No context ($\emptyset$), ConceptNet (CN), TupleKB (TP), Quasimodo (QS), GenericsKB-Best (GB), \ourkb$^{tri}$ (AT), \ourkb$^{quad}$ (AQ). AT contains only triples, while in AQ the most frequent facet in every triple is involved.}
\label{tab:qa-examples}
\end{table}

\subsection{Evaluation of Facets}

To answer RQ3, we evaluate facets both intrinsically and extrinsically.

For \textit{intrinsic evaluation}, as there are no existing CSKBs with facets, we provide comparisons with two baselines, a random permutation of facet values in \ourkb{}, and facets generated by GPT-2. First, we randomly drew 100 assertions with facets from our KB. Next, we translate each statement into a sentence prefix and ask GPT-2 to fill in the remaining words to complete the sentence. For example, given the quadruple \triple{elephant, use, their trunks, PURPOSE: to suck up water}, the sentence prefix will be ``Elephants use their trunk to'' and for this, GPT-2's continuation is ``to move around'' (see also Table~\ref{tab:facet-examples} for more examples of \ourkb{} vs. GPT-generated facets). We show each sentence prefix along with three answers (from \ourkb{}, GPT-2 and random permutation) to crowd workers and ask them to evaluate each answer along two dimensions: \textit{correctness} and \textit{informativeness}, based on a scale from 1 (lowest) to 5 (highest). Each statement is assessed by three annotators. The evaluation results are reported in Table~\ref{tab:facet-assessment}. \ourkb{} outperforms the baselines by a large margin, indicating that the facets provide valuable information to better understand the assertions. For the MTurk assessments, we obtained a mean variance score of 0.77 and a mean Pearson correlation coefficient of 0.63, indicating good agreement between annotators.

\begin{table}[t]
\centering
\scriptsize
\begin{tabular}{lp{.29\columnwidth}p{.27\columnwidth}p{.24\columnwidth}}
\toprule
\textbf{No.} & \textbf{Prefix} & \textbf{\ourkb} & \textbf{GPT-2} \\
\midrule
1 & Lawyers represent clients in & courts [location] & the case \\
\hline
2 & Elephants use their trunks to & suck up water [purpose] & move around \\
\hline
\multirow{2}{*}{3} & Artificial intelligence has a number of applications in & today's society [location] & the field of artificial intelligence \\
\hline
\multirow{2}{*}{4} & Waiters deliver food to & a table [trans-obj] & the homeless in the city of San Francisco \\
\hline
5 & Hogs roll in mud to & keep cool [purpose] & the ground \\
\hline
6 & Wine is high in & alcohol [other-qty.] & the mix \\
\bottomrule
\end{tabular}
\caption{Examples of \ourkb{}'s facet types and values along with predictions of GPT-2 given sentence prefixes.}
\label{tab:facet-examples}
\end{table}

\begin{table}[t]
    \centering
    \small
    \begin{tabular}{rrr}
        \toprule
         & \textbf{Correctness} & \textbf{Informativeness} \\
        \midrule
        Random & 1.47 & 1.29 \\
        GPT-2 & 2.85 & 2.22 \\
        \textbf{\ourkb} & \textbf{3.99} & \textbf{3.50} \\
        \bottomrule
    \end{tabular}
    \caption{Assessment of \ourkb{} and LM-generated facets.} 
    \label{tab:facet-assessment}
\end{table}

For \textit{extrinsic evaluation}, we reused the four question answering tasks from Section~\ref{subsec:extrinsic}. We incorporated facets in the context in two ways: Once based on a 256-character limit (so adding facets means that in total, fewer statements can be given as context), once by expanding the top-5-ranked statements with their facets. Note that the sets of questions in each case were different, so the absolute scores are not directly comparable. The results are shown in Table~\ref{tab:facet-extrinsic}, and the insights are twofold. On the one hand, within the fixed character-limit setting, facets do not improve results, presumably because expanding statements by facets means that some statements relevant for question answering fall out of the size limit. On the other hand, expanding a fixed number of statements by facets gives a consistent improvement in three of the four evaluation settings (FG, GG, SP), with the biggest effect being observed for informativeness in the least constrained setting (11\% relative improvement in informativeness in free generation). An example where facets are crucial is shown in Table~\ref{tab:qa-examples} with the query ``When are rats awake?''.


\begin{table}[t]
    \centering
    \small
    \begin{tabular}{lccccccc}
    \toprule
    \multicolumn{1}{l}{\multirow{2}{*}{\textbf{Context}}} & \multicolumn{2}{c}{\textbf{FG}} & \multicolumn{2}{c}{\textbf{GG}} & \multicolumn{2}{c}{\textbf{SP}} & \multicolumn{1}{c}{\textbf{MP}} \\
    \cline{2-8}
     & \textbf{C} & \textbf{I} &  \textbf{C} & \textbf{I} &  \textbf{C} & \textbf{I} & \textbf{P@5} \\
     \midrule
     \multicolumn{8}{c}{\textbf{256-character limit}} \\
     \midrule
     \textbf{\ourkb}$^{\textit{tri}}$ & \textbf{2.91} & 2.68 & \textbf{3.41} & \textbf{3.01} & 2.61 & 2.34 & \textbf{25.9} \\
    \textbf{\ourkb}$^{\textit{quad}}$ & 2.84 & 2.59 & 3.20 & 2.81 & \textbf{2.68} & \textbf{2.44} & 25.6 \\
    \midrule
    \multicolumn{8}{c}{\textbf{Top-5-statement limit}} \\
     \midrule
      \textbf{\ourkb}$^{\textit{tri}}$ & 2.73 & 2.26 & 2.91 & 2.41 & 2.20 & 1.89 & \textbf{25.8} \\
    \textbf{\ourkb}$^{\textit{quad}}$ & \textbf{2.93} & \textbf{2.53} & \textbf{3.04} & \textbf{2.57} & \textbf{2.23} & \textbf{1.96} & 25.5 \\
    \bottomrule
    \end{tabular}
    \caption{Extrinsic evaluation of facets by correctness (C) and informativeness (I).} 
    \label{tab:facet-extrinsic}
    \vspace{-0.3cm}
\end{table}

\subsection{\label{sec:per-module-eval} Per-module Evaluation}
\paragraph{Open information extraction}
We report the yield of our OIE method in comparison with StuffIE~\cite{stuffie} and Graphene~\cite{graphene} in Table~\ref{tab:openie} on a sample dataset of Wikipedia articles for ten random concepts, consisting of 2,557 sentences. Nested facets (i.e., linked contexts in Graphene) are not 
considered.
It can be seen that our extractor can identify significantly more assertions and facets than the comparison systems. Besides, the conciseness of our output improves, as average assertion length without facets (measured in words) decreases. 

\begin{table}[t]
\centering
\small
\begin{tabular}{rrrr}
\toprule
\textbf{Method} & \textbf{\#spo} & \textbf{\#facets} & \textbf{avg. length} \\ 
\midrule
{StuffIE~\cite{stuffie}} & 6,078 & 4,281 & 6.83 \\
{Graphene~\cite{graphene}} & 5,708 & 2,112 & 10.10 \\
\textbf{\ourkb} & \textbf{6,690} & \textbf{4,911} & \textbf{6.28} \\
\bottomrule
\end{tabular}
\caption{\label{tab:openie} Yield statistics of different OIE methods.}
\vspace{-0.3cm}
\end{table}

\paragraph{RoBERTa-based tasks} We report the sizes of annotated corpora and performance of our two RoBERTa classification models in Table~\ref{tab:roberta}. Since these tasks are specific to our pipeline, there are no external baselines to be compared. For both tasks, we use the pretrained RoBERTa-Base model for initialization and other specifications as follows: Adam optimizer with learning rate of $2\times10^{-5}$ and Adam epsilon of $10^{-8}$; batch size of 32; and maximal sequence length of 32. We train the model for 10 epochs for the facet labeling task, and 4 epochs for the triple pair classification task. Both models obtain very high accuracy.

\begin{table}[t]
\centering
\small
\begin{tabular}{rrrr}
\toprule
\textbf{Task} & \textbf{\#train} & \textbf{\#test} & \textbf{Acc.} \\ 
\midrule
Triple-pair classification & 21,569 & 5,392 & 0.958 \\
Facet type labeling & 3,962 & 991 & 0.928 \\
\bottomrule
\end{tabular}
\caption{\label{tab:roberta} Corpus accuracy statistics for two RoBERTa-based tasks.}
\vspace{-0.3cm}
\end{table}

%% file: sections/6-Conclusion.tex
\section{Conclusion}

This paper presented \ourkb{}, a methodology to collect advanced  commonsense knowledge about generic concepts. Our 
refined
knowledge representation allowed us to identify considerably more informative assertions, and avoid common limitations of previous works. 
The technique for
generating web search queries and filtering results
shows that CSK extraction from general web content is feasible with high precision and recall.
Intrinsic and extrinsic evaluations confirmed that the resulting CSKB is a significant advance over existing resources.


We hope that our approach revives the long-standing vision of structured CSKBs~\cite{lenat1995cyc}
and provides a cutting-edge resource that can drive forward knowledge-centric AI applications. Code, data, and a web interface are available at
\url{https://ascent.mpi-inf.mpg.de/}.
